\pdfoutput=1

\documentclass[11pt]{article}

\usepackage[]{EMNLP2022}

\usepackage{times}
\usepackage{latexsym}
\usepackage{amsfonts}
\usepackage[T1]{fontenc}

\usepackage[utf8]{inputenc}
\usepackage{graphicx}
\usepackage{tabulary}
\usepackage{booktabs}
\usepackage{amsmath}
\usepackage{multirow}
\usepackage{amssymb}

\usepackage{comment}
\usepackage{microtype}

\usepackage{inconsolata}
\usepackage{tikz}
\usepackage{adjustbox}

\title{Natural Logic-guided Autoregressive Multi-hop Document Retrieval\\ for Fact Verification}

\author{Rami Aly\\
  University of Cambridge \\
 Department of Computer Science\\
 and Technology \\
  \texttt{rami.aly@cl.cam.ac.uk} \\\And
  Andreas Vlachos \\
University of Cambridge \\
 Department of Computer Science\\
 and Technology\\
  \texttt{andreas.vlachos@cl.cam.ac.uk}
  }

\begin{document}
\maketitle
\begin{abstract}
A key component of fact
verification  is the 
evidence retrieval, often from multiple documents. 
Recent approaches use dense representations
and condition the retrieval of each document on the previously retrieved ones. The latter step is performed over all the documents in the collection, requiring storing their dense representations in an index, thus incurring a high memory footprint. 
An alternative paradigm is retrieve-and-rerank, where  documents are retrieved using methods such as BM25, their sentences are reranked, and further documents are retrieved  conditioned on these sentences, reducing the memory requirements. However, such approaches can be brittle as they rely on heuristics and assume hyperlinks between documents.
We propose a novel retrieve-and-rerank method for multi-hop retrieval,
that consists of a retriever that 
jointly scores documents in the knowledge source and sentences from previously retrieved documents using an autoregressive formulation and
is guided by a proof system based on natural logic that
dynamically terminates the retrieval process if the evidence is deemed sufficient.
This method is competitive with current state-of-the-art methods on FEVER, HoVer and FEVEROUS-S, 
while using $5$ to $10$ times less memory than competing systems.
Evaluation on an adversarial dataset indicates improved stability of our approach compared to commonly deployed threshold-based methods. Finally, the proof system helps humans predict model decisions correctly more often than using the evidence alone.

\end{abstract}

\section{Introduction}

With the growing volume of potentially misleading and false claims  \cite{graveslucasUnderstandingPromiseLimits2018}, automated fact verification \citep{Hardalov2021ASO,guo-tacl-survey-2022} is of growing interest.
A key component of open-domain fact verification systems is the retrieval of relevant documents from a knowledge base (KB) which 
provide the necessary evidence supporting or refuting a claim. Evidence retrieval accuracy correlates strongly with fact-checking accuracy, 
as observed in a recent shared task \cite{aly-etal-2021-fact}.

\begin{figure}
            \fbox{\begin{minipage}{19em}
                    \small    
                    \textbf{Claim:} The 66th Primetime Emmy Awards was hosted by an Iraqi comedian born in 1973.
                    \rule{\linewidth}{0.1em}
                    \raggedright{\textbf{Evidence Documents:}} \\
                    \vspace{0.5em}
                    \colorbox{yellow}{\underline{66th Primetime Emmy Awards}}\\
                    The 66th Primetime Emmy Awards honored the best in U.S. prime time television programming from June 1, 2013 until May 31, 2014, as chosen by the Academy of Television Arts \& Sciences. 
                    Comedian and Late Night host \textcolor{red}{Seth Meyers hosted the ceremony} for the first time.
                    \vspace{0.5em}
                    \colorbox{yellow}{\underline{Seth Meyers}} \\
                    Seth Adam Meyers (\textcolor{red}{born December 28 , 1973}) is an \textcolor{red}{American comedian} , writer , producer , political commentator , actor , media critic , and television host. He hosts Late Night with Seth Meyers, a late-night talk show on NBC. Prior to that, he was a cast member and head writer for NBC's Saturday Night Live (2001–2014). 
                    \rule{\linewidth}{0.1em}
                    \textbf{Verdict:} Refuted
            \end{minipage}}
        \caption{A FEVER example where multiple documents are required for verification (relevant evidence in red). 
        }
        \label{fig:feverous_example}
    \end{figure}

\begin{figure*}[h]
	\centering
	\includegraphics[width=0.97\textwidth]{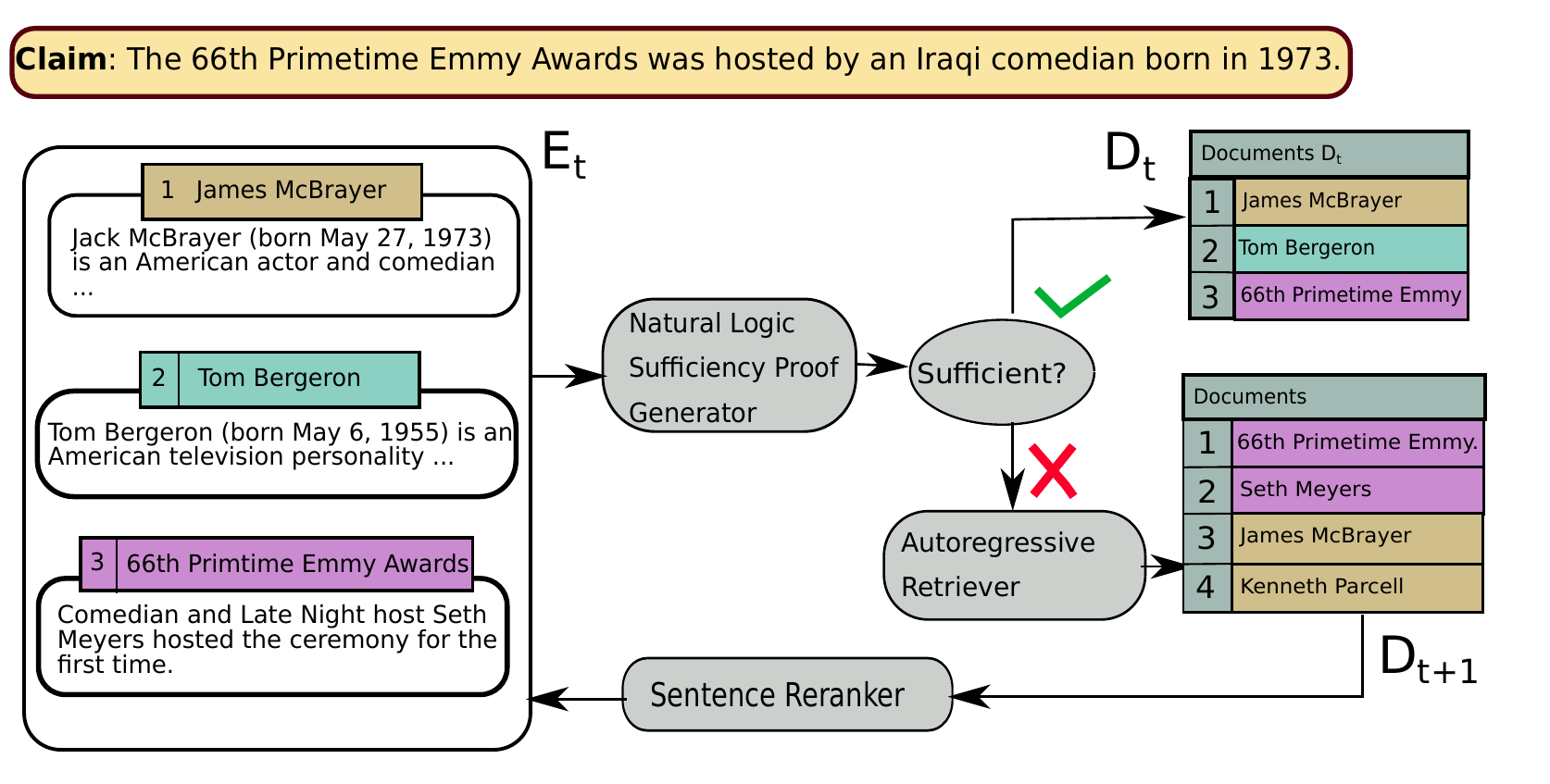}
	\caption{The AdMIRaL pipeline. At hop $t$, given the claim and sentences $E_{t}$ from documents $D_t$, a proof is generated to predict whether the evidence $E_{t}$ is sufficient for verification or whether additional evidence is needed. If sufficient, the retriever terminates, otherwise, our autoregressive retriever scores documents in the KB jointly with $E_t$ updating the documents to $D_{t+1}$, before they are passed to the sentence reranker to obtain $E_{t+1}$. }
	\label{fig:method}
\end{figure*}

Document retrieval for fact verification can be complex, as required evidence is often found in multiple documents, with each document containing partial information needed to assess the veracity of a claim. An example is shown in Figure~\ref{fig:feverous_example}. Given the claim ``\emph{The 66th Primetime Emmy Awards was hosted by an Iraqi comedian born in 1973.}", information from two documents has to be considered to verify the claim: one mentions the person who hosted the Awards, and another provides information about this person. The claim itself is often not leading to the second document, and it is rather misleading as it mentions incorrect information that would retrieve irrelevant documents (i.e.\ about Iraqi comedians). 
Instead we need to condition the retrieval of some evidence pieces on other evidence documents, (i.e.\ that \emph{Seth Meyers} is the host of the \emph{66th Primetime Emmy Awards}). 


Recent approaches to multi-hop retrieval for fact verification commonly use dense representations for both claim and documents, 
and condition the retrieval of each document on the previously retrieved ones 
\citep{xiongAnsweringComplexOpenDomain2020, khattabBaleenRobustMultiHop2021}. After each iteration (hop), the claim representation is modified and compared against the entire KB, necessitating to store all 
document representations in a dense index. 
 However, typically millions of documents are considered, thus the dense index has a large memory footprint
(see Table~\ref{tab:memory-req-dense}). 
An alternative paradigm is Retrieve and Rerank (RnR) \citep{Nie2019CombiningFE, stammbach-2021-evidence, malon-2021-team}, where candidate documents are retrieved, sentences from them are reranked, and then conditioned on the top-ranked sentences additional documents are retrieved. By 
retrieving the candidate documents
using sparse retrievers (e.g.\ BM25) a dense index becomes unnecessary, while a dense reranker can still be used to take advantage of dense representations. Yet, to reach competitive multi-hop performance, RnR systems assume links between documents and rely on heuristics, such as down-weighting hyperlinked documents by a fixed factor, assuming evidence from the first iteration is more important. Heuristics might not generalize well across datasets, and while links between documents are beneficial when available (e.g.\ in Wikipedia), many textual KBs do not have them. 

\begin{table}[ht!]
	\centering
	\resizebox{1\linewidth}{!}{
	\begin{tabular}{l| ccc }
		\toprule
		& \multicolumn{2}{c}{Datasets}\\
		Model & HoVer & FEVER & FEVEROUS-S\\
		\midrule
		BM25 & 3.0GB &  6.1GB & 20.4GB\\
		\midrule
		MDR & 32.1GB & 72.8GB & -- \\
		ColBERT & 81.3GB & -- & --\\
		ColBERTv2 & 16.0GB & 34.4GB  & 124.2GB\\
		\hline
		AdMIRaL (Ours) & 3.3GB & 6.4GB & 20.7GB \\
		\bottomrule
	\end{tabular}}
	\caption{
	Memory footprint for different datasets of several sparse/dense retrieval models. 
	FEVER/HoVer consider only Wikipedia introductions, while FEVEROUS-S consists of entire Wikipedia pages. }
	\label{tab:memory-req-dense}
\end{table}

To address these challenges we propose AdMIRaL (\underline{A}utoregressive \underline{d}ocument \underline{M}ulti-hop \underline{I}nformation \underline{R}etrieval with N\underline{a}tural \underline{L}ogic-guidance), 
a novel multi-hop document retriever for RnR that consists of two components: i) a retriever that jointly scores documents in the KB and sentences reranked from previously retrieved documents 
using an autoregressive formulation \citep{decaoAutoregressiveEntityRetrieval2021}, 
ii) a proof system using Natural Logic \cite{maccartney2014natural} 
to 
assess the sufficiency of the evidence retrieved to verify a given claim, and terminate the retrieval 
of further documents. 
The method is illustrated in Figure~\ref{fig:method}. 
By retrieving using an autoregressive formulation, 
generating document and sentence identifiers jointly token by token and conditioned only on the context, AdMIRaL does not need to store dense representations in an index. 
The proof system controls the merging of evidence documents between hops while being faithful and interpretable with regards to system's operation in each hop.

We improve document recall and F$_1$ by 1.4\% and 4.6\% over the state-of-the-art performance on FEVER \citep{thorneFEVERLargescaleDataset2018} respectively, 
and are competitive with state-of-the-art performance on HoVer \citep{jiang2020hover} and the sentence-only version of FEVEROUS (i.e.\ excluding tables) \citep{alyFEVEROUSFactExtraction2021}, while using $5$ to $10$ times less memory than competing dense retrieval systems and a runtime complexity more favourable when scaling to large KBs. 
We further assess the robustness of AdMIRaL on an adverserial version of FEVER \citep{Hidey2020DeSePtionDS}, and show performance gains using various initial retrievers. 
Finally, human evaluation indicates that the natural logic proofs help humans predict model decisions correctly more often than using the evidence directly.\footnote{\url{https://github.com/Raldir/AdMIRaL}} 

\section{Related Work}

Early approaches to multi-hop document retrieval for automated fact verification are based on the RnR paradigm. They use sparse or entity-linking based retrievers to find candidate documents (e.g.\ \citep{hanselowski-etal-2018-ukp}), rerank sentences in a classification formulation (such as ESIM \citep{thorneFEVERLargescaleDataset2018} or pre-trained encoders \citep{liuFinegrainedFactVerification2020, Zhong2020ReasoningOS}), and use hyperlinks to find additional documents  \citep{Nie2019CombiningFE, stammbach-neumann-2019-team}. The aforementioned approaches are limited to two iterations, using hyperlinks
extracted from the initial list of candidate sentences to be considered as additional documents in a second iteration. Assuming that the evidence from the first iteration (i.e.\ initial retrieval) is more important, \citet{stammbach-2021-evidence} down-weight hyperlinked documents by a fixed factor.
\citet{malon-2021-team} proposes the use of a generative model that imagines missing evidence sentences and selects new sentences based on word overlap. 
Indirect improvements to multi-hop RnR are achieved through stronger document retrieval models, such as GENRE \citep{decaoAutoregressiveEntityRetrieval2021}, or more accurate rerankers \citep{stammbach-2021-evidence, jiang-etal-2021-exploring-listwise}. GENRE in particular produced state-of-the-art results on document retrieval for fact verification, by generating documents using an autoregressive formulation, not necessitating a dense index. 

Multi-hop dense passage retrieval (MDR) \citep{xiongAnsweringComplexOpenDomain2020} iteratively retrieves evidence documents using dense passage retrieval (DPR) \citep{Karpukhin2020DensePR}, a bi-encoder that encodes claim and documents separately and uses efficient maximum inner-product search to score each document in the KB. 
Since the search space grows exponentially with each iteration in the number of documents in the KB, MDR uses beam search to aggressively prune the search space which reduces scalability to many hops. 
In contast, \citet{khattabBaleenRobustMultiHop2021}'s Baleen retriever performs multi-hop document retrieval by condensing retrieved documents after each iteration into a condensed context (i.e. sentence(s)) that is used to update the dense representation of the claim, reducing the search space by omitting all other candidates. While condensing is similar to the reranking step of RnR, Baleen then still scores each document in the KB at each hop, like MDR. They further propose late interaction (FLIPR), 
which allows different parts of the claim to match different relevant parts of documents. 
Indicative of the challenge of scaling to large KBs is that
out of $12$ systems submitted to the FEVEROUS shared task \citep{aly-etal-2021-fact}, with FEVEROUS being the fact verification dataset with the largest KB) not a single one opted to use a dense retrieval index.

Beyond fact verification, in question answering (QA), sophisticated graph based retrieval models have been proposed that make explicit use of links in the KB \citep{Asai2020LearningTR, li2021hopretriever}. 
Of particular relevance is the approach of
\citep{qi-etal-2021-answering}, that uses the feedback of the reader after each iteration to determine whether the answer has been retrieved and is sufficient, thus determining the number of hops dynamically instead of fixing it in advance. 
In their QA benchmark, all answers to the questions can be found in the source; however in many fact verification datasets a key challenge included are claims that cannot be verified using the KB. Moreover, as explained false information mentioned in claims can be misleading, 
making the stopping criterion based on sufficiency for fact checking more challenging. 

\section{AdMIRaL}


Given a KB consisting of documents $\mathbb{D}$, the task of document retrieval for fact verification is to find the set documents $D=\{d_1, \ldots, d_n\} \subset \mathbb{D}$ that is sufficient to support or refute a claim $c$.\footnote{In FEVER, if a claim is labelled with not enough evidence (\textit{NEI}) it has no documents associated with it, unlike in FEVEROUS and HoVer (HoVer merges \textit{refuted} and \textit{NEI} instances in one class, \textit{not supported}).} We assume that each document is associated with a unique document title, following previous work \citep{decaoAutoregressiveEntityRetrieval2021}. In the case of multi-hop retrieval $n$ must be larger than 1. 
 In the RnR paradigm 
 the retrieval is done in three steps: (i) find and return the $k$ best-scoring document sets $D_t = \{D^1_t, \ldots D^i_t \dots D^k_t\}$, with $t$ being the current retrieval iteration, with $t\geq1$ and $|D^i_t| = t$, and $D^i_1$ the $i$th best-scoring single-hop document set of length $1$ (i.e. a single document), (ii) rerank sentences from the $k$ document sets in $D_t$ into the top $l$ sentences $E_t$, 
 iii) use $E_t$ to update the set of document sequences to D$_{t + 1}$. Steps two and three are then repeated for a total of $n$ hops. Since $n$ is not known a priori for a given claim previous work sets $n$ to an upper bound of the dataset.
 
 \begin{figure*}[h]
	\centering
	\includegraphics[width=1\textwidth]{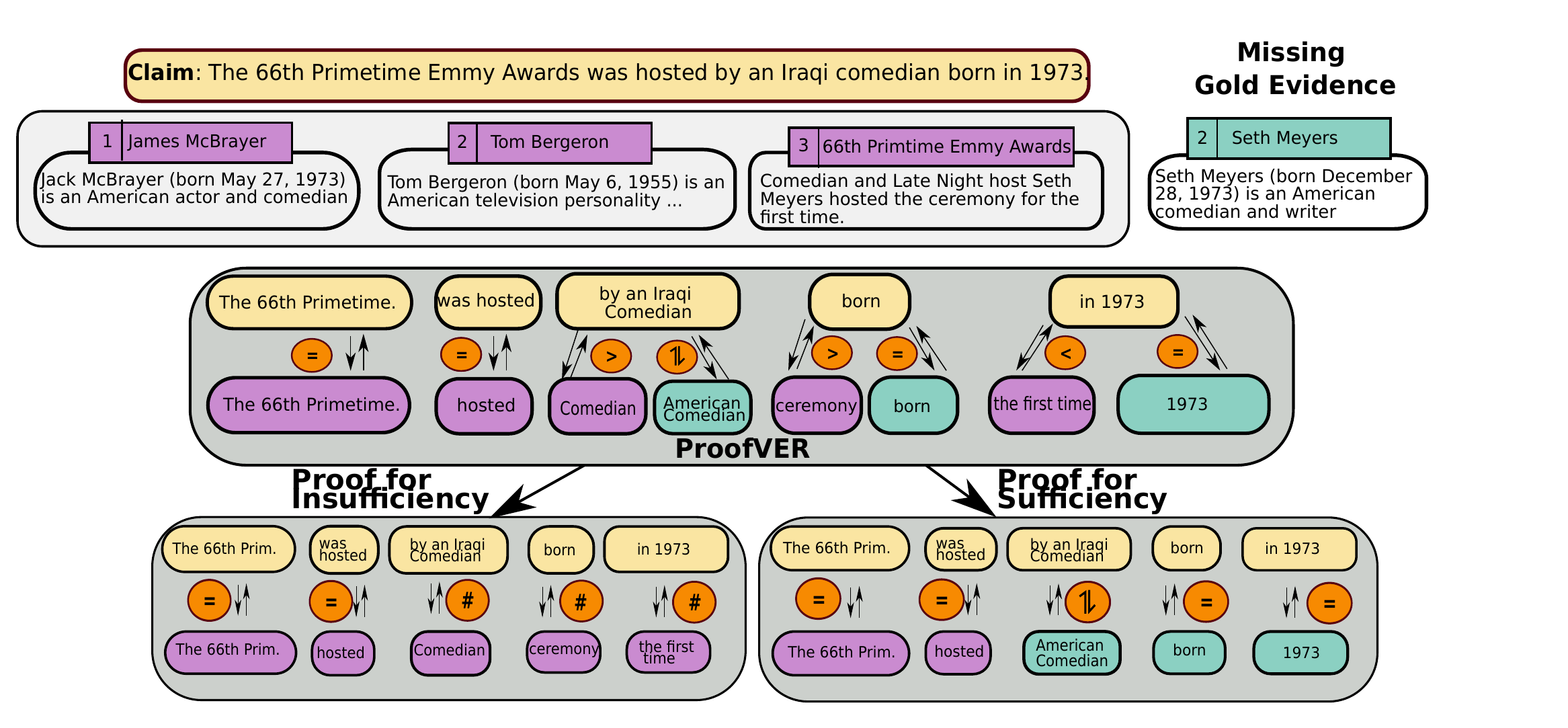}
	\caption{Left: Illustration of the process of generating sufficiency proofs for training. For a given claim two proofs are generated: one given sufficient and insufficient evidence. Insufficiency is predicted iff any mutation in the proof sequence is assigned the independence operator. 
	}
	\label{fig:sufficiency_proof}
\end{figure*}

 AdMIRaL performs the third step of RnR in two stages: (i) the retrieval of D$_{t+1}$ by jointly scoring $\mathbb{D}$ and $E_t$ 
 using a generative model in an autoregressive framework that cross-attends over the claim $c$ and $E_t$, (ii) a dynamic retrieval termination criterion formulated as an evidence sufficiency task by generating a proof of Natural Logic \cite{maccartney2014natural} based on Proofver \cite{krishna2021proofver}, hence the number of hops $n_{dyn}$ for AdMIRaL is being determined dynamically for each claim given the evidence $E_t$, with $n_{dyn} \leq n$.
 
 \subsection{Autoregressive Document Retrieval}
 Given a claim $c$  and the top-ranked sentences E$_t$, 
 we formulate AdMIRaL as a pointwise reranker for document sets $D_{t+1}^i$ of length $t+1$, i.e. sets containing one more document (hop) than the ones in D$_t$. The scoring function for $D_{t+1}^i$ is defined jointly over $D_{t}^i$ and $d_{t+1}$. We consider the sentences $E_t$ to be an approximation of all relevant information in $D_t$ regarding $c$, yet in much more compact form. Hence, we define the score of a set of documents
 $D_{t}^i$ as the sum of the scores of all its possible underlying sentence combinations :
 \begin{align}
&\text{score}(D_{t+1}^i| c, E_t) \nonumber\\
&=\text{score$_d$}(d_1, \ldots, d_t, d_{t+1}| c, E_t) \nonumber \\
&= \sum_{s_1\in d_1, \ldots, s_t\in d_t} \text{score$_s$}(s_1, \ldots s_t, d_{t+1}| c, E_t),
\label{eq:1}
 \end{align}
 with $d_{t+1} \in \mathbb{D}$. Thus Eq.\ref{eq:1} scores $D_{t+1}^i$ jointly on $\mathbb{D}$ and $E_t$. 
Since we assume all relevant information of $D_t$ is in $E_t$, the scoring function $\text{score$_s$}(\mathcal{S}, d_{t+1}| c, E_t)$, with $\mathcal{S} = \{s_1, \ldots , s_t\}$, only scores sets $\{\mathcal{S}, d_{t+1} \}$ where all sentences of $\mathcal{S}$ are in $E_t$. 
The scoring function is computed using a generative model with an autoregressive formulation over the unique document titles, conditioning the score of $d_{t+1}$ on $\mathcal{S}$:
 \begin{align}
& \text{score}(\mathcal{S}, d_{t+1}| c, E_t)\nonumber \\
& = p_{\theta}(\mathcal{S}|c, E_t) \cdot p_{\theta}(d_{t+1}| c, E_t, \mathcal{S}) \nonumber\\
&  = 
\underbrace{\prod_{u=1}^{M}p_{\theta}(q_u |q_{<u}, c, E_t)\nonumber}_{p_{\theta}(\mathcal{S}|c, E_t)}\\
&\cdot \underbrace{\prod_{m=1}^{N}p_{\theta}(y_m |y_{<m}, c, E_t, \mathcal{S})}_{p_{\theta}(d_{t+1}| c, E_t, \mathcal{S})},
\label{eq:2}
\end{align}
, where $\mathbf{y}$ is the sequence of tokens representing the title of document $d_{t+1}$, $\mathbf{q}$ is the sequence of tokens representing the sentence identifiers (i.e.\ unique sentence ids encoded in $E_t$) of $\mathcal{S}$, and $\theta$ are the parameters of the model. 

The scoring model is a generative pre-trained transformer-based architecture, namely BART \cite{lewis-etal-2020-bart}, allowing us to cross-encode claim $c$ and sentences $E_t$ while using the model's language understanding and knowledge capabilities
\citep{petroni-etal-2019-language, radford2019language}. Since the document sequences are scored using only the claim and the information $E_t$, no document representations have to be pre-computed and stored in an index. 
However, since $|\mathbb{D}|$ is very large, it is infeasible to compute a score for each set $\{\mathcal{S}, d_{t+1} \}$, instead we use beam search to efficiently navigate the search space, searching only for the $q$ top-ranked sequences. Note that the beam search for generation differs substantially from the one used for iterative retrieval by \citet{xiongAnsweringComplexOpenDomain2020}: our search is over the model's vocabulary and using a softmax operation for scoring, not over the entire KB with a MIPS comparison between all dense representations, the former being substantially more efficient with a much smaller search space. 
Since in traditional decoding any token from the vocabulary can be generated at any position, we might generate sequences that are non-existing document/sentence identifiers. We follow  \citet{decaoAutoregressiveEntityRetrieval2021} by constraining the generation using a prefix tree (trie). 
In practice, we have to switch between two search spaces: the sentence and document identifiers, as we want to ensure that sentence identifiers are generated first to condition the generation of $d$. We achieve this by employing dynamically constrained markup decoding \citep{decaoAutoregressiveEntityRetrieval2021}, where markups are used during encoding to switch between search spaces. The $q$ top-ranked document sets are then returned as $D_{t+1}$.

 \paragraph{Training} We train a separate model for each hop $t$ using maximum likelihood estimation,
 following the Neural Machine Translation fine-tuning of BART \cite{lewis-etal-2020-bart}, computing the log probability of a document title (as a sequence of tokens) given $E_g$ and the claim $c$. Given an ordered list of gold evidence sentences $E_g$ of length $m$, with $m \leq n$, we consider as input during training the first $m-1$ evidence elements, and the document title of the remaining evidence sentence to be $y$. In cases where an explicit ordering of evidence sentences is not available, we generate all $t-1$ combinations of splitting $E_g$ into input and output data. See appendix \ref{app:autoregress} for details.

\subsection{Evidence Sufficiency with Natural Logic}

To determine whether the retrieved evidence is sufficient for the claim being verified, 
we generate a proof using natural logic \cite{maccartney2014natural}, inspired by recent work who used it for verification \cite{krishna2021proofver}.
Given a claim $c$ and the evidence sentences $E_t$, a seq2seq model generates a proof sequentially in an autoregressive formulation, from left to right.  Each part of the claim $c$ is sequentially mutated into an evidence span of $E_t$, with a natural logic operation (NatOp) defining the nature of the mutation. We consider four out of seven NatOps defined in \citep{maccartney2014natural} for the sufficiency proof: equivalence ($\equiv$), negation ($\neg$), alternation ($\downharpoonleft \! \upharpoonright$), and independence (\#).  See Figure \ref{fig:sufficiency_proof} for an example (bottom left). Mutations between semantically equivalent spans are assigned the equivalence NatOp ($\equiv$), such as \emph{The 66th Primetime Emmy Awards}. The mutation of the claim span \emph{by an Iraqi comedian} is assigned the independence NatOp (\#), indicating that no related evidence span exist in the Evidence $E_t$. 
We do not consider the cover NatOp ($\smallsmile$), forward entailment NatOp ($\sqsubseteq$) and reverse entailment NatOp ($\sqsupseteq$) as they are not conclusive indicators for sufficiency and can be replaced with independence for our purposes. For instance, Africa  $\sqsubseteq$ Tunisia holds, yet given a claim ``Ryan Gosling has been to Africa." and evidence ``Ryan Gosling has been to Tunisia.", further evidence that links Tunisia to Africa is required for the evidence to be sufficient. 





To determine the sufficiency based on the generated proof, we consider the sequence of operators assigned to each mutation. We predict insufficiency iff any mutation in the proof sequence is assigned the independence operator. The proof-based sufficiency prediction is faithful by construction and provides an and explainable sufficiency prediction for multi-hop systems.

Since claim-evidence pairs annotated with natural logic proofs for sufficiency prediction are not  available, we generate them. For each claim we generate two proofs: one based on insufficient and one on sufficient evidence (see Figure \ref{fig:sufficiency_proof}). Given incomplete and complete evidence for a claim, we first use ProofVER \cite{krishna2021proofver} to generate the respective initial proofs. However, since ProofVER has been trained on data annotated with heuristics targeting the assessment of a claims veracity, ProofVER's NatOps specifying  mutations are not suitable for our purpose of predicting the sufficiency of evidence. 
Hence, we reassign NatOps in each initial proof to ensure its consistency with the sufficiency of the input evidence. First, all forward/reverse entailment NatOps ($\sqsubseteq$)/($\sqsupseteq$)  are replaced with independence NatOps (\#). 
We then modify the fewest NatOPs in a proof possible to reach the correct (in-)sufficiency prediction. For proofs that indicate sufficiency of evidence which is insufficient, we assign the independence NatOp  to the mutation with the most dissimilar claim and evidence spans, measured using cosine similarity of the mean-pooled contextual representation of a pre-trained language model. 
If a proof indicates insufficiency but is sufficient, we first search for claim/evidence terms in multiple lexicons (Wordnet \citep{miller-1995-wordnet}, Synonym Antonym pairs of \citep{roth-schulte-im-walde-2014-combining}, and PPDB \citep{pavlick-etal-2015-ppdb}) to find suitable matches, and then replace all remaining independence NatOps with equivalence if the claim is supported, or negation if refuted.

\section{Experimental setup}

\begin{table*}[ht!]
	\centering
	\resizebox{0.82\linewidth}{!}{
	\begin{tabular}{ll| cc cc cc}
		\toprule
	&	& \multicolumn{4}{c}{Recall@5} & \multicolumn{2}{c}{Recall@100} \\
	&	Model/Datasets & \multicolumn{2}{c}{FEVER} & \multicolumn{2}{c}{FEVEROUS-S} & \multicolumn{2}{c}{HoVer}\\
	& & 2-hop & Overall & 2-hop & Overall & 2-hop & Overall \\
		\midrule
	\multirow{ 4}{*}{Single-Hop} &	BM25  & 0.150  & 0.658 &  0.410 & 0.752 & 0.789 & 0.397 \\
	 & GENRE & 0.191  & 0.892 & 0.330 & 0.705 & 0.382 & 0.107 \\
	 & AdMIRaL single-hop  & 0.357 & 0.928 & 0.441 & 0.799 & 0.886 & 0.470  \\
	 & DPR  & 0.191 & 0.754 & -- & -- & --& -- \\
	 \hline
	\multirow{ 4}{*}{Multi-Hop} 
	& Hyperlinks & 0.667 & \underline{0.945} & \underline{0.506} & \underline{0.822} & 0.904 & 0.641 \\
	&	MDR$^\dagger$ & \underline{0.691} & -- & --&-- & -- & -- \\
	&	ColBERT-Hop$^*$ & -- & -- & --&-- &  \underline{0.958} &  0.748\\
	&	Baleen$^*$ & -- & -- & --&-- & \textbf{0.977} & \textbf{0.922} \\
		\hline
	&	AdMIRaL (Ours) & \textbf{0.705} & \textbf{0.956} & \textbf{0.610} & \textbf{0.847} & \textbf{0.977}  & \underline{0.817} \\
		\bottomrule
	\end{tabular}}
	\caption{Document retrieval scores for 2-hop, and overall scores. 
	To compare with previous work on HoVer, we report recall@100 for \emph{supported} claims on dev. $\dagger$ and $^*$ indicate results taken from \citet{xiongAnsweringComplexOpenDomain2020} and \citet{khattabBaleenRobustMultiHop2021}, respectively. \textbf{Bold} numbers indicate best and \underline{underline} the second-best score.
	}\label{tab:overview-document-retrieval}
\end{table*}

\paragraph{Datasets}
We evaluate our multi-hop document retriever on FEVER \citep{thorneFEVERLargescaleDataset2018}, FEVEROUS \citep{alyFEVEROUSFactExtraction2021}, and HoVer \citep{jiang2020hover}. FEVER consists of claims that predominantly require a single evidence sentence (87\%). Contrary to \citet{xiongAnsweringComplexOpenDomain2020} who evaluate only on the multi-hop part on FEVER, we report results for multi-hop and the entire dataset; this is more realistic as in practice it is unknown in advance whether a claim requires multi-hop document retrieval. HoVer contains of 46\%, 36\%, and 18\% two-hop, three-hop, and four-hop claims, respectively. Contrary to FEVER, and HoVer that only consider the introductory section of Wikipedia pages, FEVEROUS considers entire Wikipedia articles, including semi-structured evidence in the form of table (cells) and contains of 16\% multi-hop claims. From FEVEROUS we consider only the claims that require exclusively sentence evidence, as we focus on text retrieval (FEVEROUS-S), which constitutes about 41\% of claims in FEVEROUS.


\paragraph{Implementation Details}
The autoregressive model for both retrieval and proof generation is BART \citep{lewis-etal-2020-bart}, which are trained independently from each other. 
For the initial retrieval, we first retrieve document candidates $D_{1}$ for all three datasets using GENRE, fine-tuned on the KILT version of FEVER \citep{petroni-etal-2021-kilt}, and BM25 based on Pyserini \citep{lin2021pyserini}. To rerank the sentences of these documents and keeping the top $l=5$ in $E_1$, we use the token-level evidence selection model of \citep{stammbach-2021-evidence} for FEVER, and the pointwise T5 reranker of \citep{jiang-etal-2021-exploring-listwise} for HoVer, and FEVEROUS-S. For FEVER and FEVEROUS-S we consider the top $10$ documents for reranking while for HOVER we focus on the top $100$, to keep scores comparable to previous work.

\section{Results}

\subsection{Multi-hop Document Retrieval}
Document retrieval results on each dataset's dev set are shown in Table \ref{tab:overview-document-retrieval}.  Results include single-hop retrievers, covering sparse retrieval (BM25), entity-based (GENRE), and dense passage retrieval (DPR). We further show the scores of the single-hop retriever used for AdMIRaL, namely AdMIRaL single-hop. We further compare AdMIRaL against state-of-the-art multi-hop retrieval approaches, including MDR, Baleen, and ColBERT-Hop \citep{khattabBaleenRobustMultiHop2021}, which all necessitate a dense index. We further compare against an RnR retriever that makes explicit use of hyperlinks in sentences to retrieve new documents, as done in the multi-hop setting of \citet{stammbach-2021-evidence}. The memory and computational requirements to run dense retrievers on FEVEROUS-S exceeded our resources (see Appendix \ref{app:impl}), hence results on FEVEROUS-S are only computed for RnR, i.e.\ \citet{jiang-etal-2021-exploring-listwise} with hyperlinks, and single-hop approaches.
As seen in Table \ref{tab:overview-document-retrieval}, AdMIRaL achieves the highest 2-hop recall score on all datasets, and the highest overall recall on FEVER and FEVEROUS-S, falling only behind Baleen on HoVer. AdMIRaL improves 2-hop recall of the initial retrieval by 34.8\% percentage points on FEVER, 16.9\% on FEVEROUS-S and by 9.1\% on HoVer. For the latter, overall scores increased by even larger 34.7\%, as AdMIRaL improves retrieval substantially for HoVer claims requiring more than 2 hops. Furthermore, AdMIRaL is more precise than state-of-the-art models, achieving an F$_1$ improvement over MDR on FEVER by 4.6\% and over hyperlinks on FEVEROUS-S by 14.2\% (see Appendix \ref{app:further-results}).

\subsection{Efficiency}
\paragraph{Memory Footprint} AdMIRaL achieves overall competitive performance while being an order of magnitude more space efficient. AdMIRaL's memory footprint is composed of the inverted index for the initial BM25 retrieval and the prefix tree of the document titles (excluding the model itself), resulting in a footprint of 3.3GB, 6.4GB and 20.7GB for HoVer, FEVER, and FEVEROUS-S, respectively. This is about $10$ times less than MDR, $5$ times less than Baleen (ColBERTv2), and $27$ times less than ColBERT (see Table \ref{tab:memory-req-dense}), since these necessitate a dense index of all documents in the KB. The footprint of AdMIRaL is comparable to RnR approaches \cite{stammbach-2021-evidence}, as the inverted index is only a few hundred Megabytes of size, negligible to the size of the inverted index.

\paragraph{Runtime Efficiency} The runtime complexity of a retriever consists of two components: step (i) the indexing of the KB and (ii) the retrieval itself.\footnote{The indexing efficiency is relevant as a KB’s content frequently updates in the real world and so then must the index.} Since AdMiRaL only builds a BM25 index for the initial retrieval - step (i) - is substantially faster than for dense retrievers such as Baleen. Specifically, it takes 2 minutes and 32s to build the HoVer index for AdMIRaL (and for \citeauthor{stammbach-2021-evidence}'s RnR retriever), compared to 290 minutes and 12s for Baleen’s dense index. For step (ii), AdMIRaL scales better than dense retrievers with respect to both KB size and the number of iterations. Dense retrievers such as Baleen or MDR scale by $\mathcal{O}(n  |\mathbb{D}|)$, with  $|\mathbb{D}|$ being the number of documents in the KB, as they do a comparison to all documents in the KB at every iteration for a total of $n$ iterations, with $n$ being the number of iterations set to an upper bound. In contrast, for AdMIRaL, only the initial retrieval depends on the size of the KB (i.e. BM25), while the autoregressive retrieval at each hop depends on the model’s vocabulary size, hence $\mathcal{O}(n_{dyn} + |\mathbb{D}|)$ (including initial retrieval, otherwise $\mathcal{O}(n_{dyn})$), with $n_{dyn} \leq n$ being the number of iterations according to the dynamic termination of AdMIRaL. \footnote{Reducing runtime through dynamic termination with AdMIRaL would be challenging as the computational cost of the proof generator itself is at least as high as the autoregressive retriever. Investigating how dynamic termination can improve multi-hop retrieval efficiency is an interesting future direction to explore but is outside of our focus of AdMIRaL.} However, the underlying constant of AdMIRaL is large as it relies on two Encoder-Decoder models (autoregressive retrieval + proof generation) and a sentence reranker which makes it computationally more expensive when used on relatively small KB’s such as HoVer. We measure 2.87s on average for a single HoVer query on AdMIRaL, 1.94s for Baleen, and 0.69s for \citet{stammbach-2021-evidence}. However, on large KBs (such as FEVEROUS-S with 7x the size of HoVer), the runtime is more favourable for AdMIRaL than Baleen. While the memory requirements to run Baleen on FEVEROUS-S exceed our resources, already on a KB with twice the size of HoVer, Baleen takes 2.30s, while AdMiRaL's runtime is nearly unchanged (2.88s).

\section{Discussion}

\paragraph{Autoregressive Generation}
We compare the auto-regressive document scoring method of AdMIRaL to some variants, namely a model that i) considers only the top sentence of $E_t$ for scoring and generation (Top-1) ii)  does not score documents and sentences jointly, instead it only scores documents, and the top-ranking documents are concatenated to the $t$ highest-ranked documents of $D_t$ (Not-joint) iii) scores a ranked set of documents directly (i.e.\ scoring $\mathbb{D}^t$) (Joint-docs). We also evaluate a AdMIRaL model that exploits hyperlink information by concatenating a sentence's hyperlinks to its the end before being passed as input. 
Results are shown in Table \ref{tab:autoregressive-generation-queries}. While \emph{Top-1} achieves a comparable exact match score to AdMIRaL (even slightly higher for two-hop claims), its recall is considerably lower. On the contrary, \emph{Not-joint} achieves competitive recall, yet, lags behind in terms of exact match accuracy, as the original order of the top-ranked documents in $D_t$ are largely unchanged.  Finally, \emph{Joint-docs} performs worst overall, likely due to the difficulty of evidence ordering during training, as also observed by \citep{xiongAnsweringComplexOpenDomain2020}. Incorporating hyperlink information into AdMIRaL improves recall substantially, resulting to a FEVER multi-hop state-of-the-art improvement of $0.16$ percentage points.
\begin{table}[ht!]
	\centering
	\resizebox{1\linewidth}{!}{
	\begin{tabular}{l| cccc }
		\toprule
		 & \multicolumn{4}{c}{FEVER}\\
		 Model & \multicolumn{2}{c}{R@5} & \multicolumn{2}{c}{EM}\\
		 & Two-hop & Overall & Two-hop & Overall  \\
		\midrule
	  Initial & 0.35 & 0.93 & 0.26  & 0.87 \\
		Top-1 & 0.67 & 0.94 & 0.52 & 0.88  \\
		Not-Joint & 0.71 & 0.96 & 0.32 & 0.87\\
		Joint-docs & 0.59 & 0.94 & 0.43 & 0.87   \\
		\hline
		Ours & 0.71 & 0.96 & 0.51 & 0.89  \\
		Ours w/ hyperlinks & \textbf{0.85} & \textbf{0.97} & \textbf{0.51} & \textbf{0.89} \\ 
		\bottomrule
	\end{tabular}}
	\caption{Document retrieval scores for several variations to proposed autoregressive retriever,  R@5: Recall@5, EM: Exact Match Accuracy. 
	}\label{tab:autoregressive-generation-queries}
\end{table}

\paragraph{Robustness}
We further evaluate the robustness of our model by evaluating AdMIRaL on an adversarial fact verification dataset DeSePtion \citep{Hidey2020DeSePtionDS}, which consists adversarial attacks generated as part of the FEVER2.0 adversarial shared task \citep{thorne-etal-2019-fever2}. The attacks consider lexical variations/substitutions, entity disambiguation, (multi-hop) temporal reasoning, multiple prepositions and multi-hop reasoning. Document level results for models trained on FEVER and evaluated on DeSePtion are shown in Table \ref{tab:adversarial-fever}. AdMIRaL achieves substantial increases over the initial retriever and a BM25 baseline. Moreover, while \emph{Not-joint} achieves similar recall to AdMIRaL on FEVER, it performs worse on the adversarial dataset. This highlights the brittleness of adding new documents statically to the top-ranking documents using a fixed position or threshold, as commonly done. 
\begin{table}[ht!]
	\centering
	\resizebox{1\linewidth}{!}{
	\begin{tabular}{ll| cccc }
		\toprule
		& & \multicolumn{4}{c}{Datasets}\\
		& Model & \multicolumn{2}{c}{FEVER} & \multicolumn{2}{c}{Adversarial-FEVER}\\
		& & Two-hop & Overall & Two-hop & Overall  \\
		\midrule
		\multirow{3}{*}{R}
    	& 	Initial & 0.36 & 0.93 & 0.54  & 0.77 \\
		& BM25 & 0.15  & 0.658  & 0.22 &  0.55 \\
		& Not-Joint  & \textbf{0.71} & \textbf{0.96} & 0.72 & 0.84 \\
		& Ours & \textbf{0.71} & \textbf{0.96}  & \textbf{0.74} & \textbf{0.86}  \\
		
		\hline
		\multirow{3}{*}{EM}
		& Initial & 0.26 & 0.87 & 0.48  & 0.69 \\
		& BM25 & 0.06 &  0.40 & 0.13 &  0.36 \\
		& Not-Joint  & 0.32 & 0.87 & 0.50 & 0.71 \\
		& Ours & \textbf{0.51} & \textbf{0.89} & \textbf{0.56} & \textbf{0.73}  \\
		\bottomrule
		\bottomrule
	\end{tabular}}
	\caption{Document retrieval scores on the adversarial dataset, R: Recall@5, EM: Exact Match accuracy.
	}\label{tab:adversarial-fever}
\end{table}

\paragraph{Results with different initial retrievers} Another aspect we analyze is the stability of AdMIRaL with different initial retrievers, i.e.\ the retrieval of $E_1$. Results are shown in Table \ref{tab:initial-retriever}. The relative improvements achieved by AdMIRaL are consistent across retrievers, improving recall@5 for BM25, KGAT \citep{liuFinegrainedFactVerification2020}, and \citep{jiangExploringListwiseEvidence2021} (and \citep{stammbach-2021-evidence} as used in AdMIRaL) on FEVER, by an average of 33\% percentage points with a variance of $0.0004$.

\begin{table}[ht!]
	\centering
	\resizebox{1\linewidth}{!}{
	\begin{tabular}{l| cc cc }
		\toprule
		 Initial Retriever & \multicolumn{2}{c}{Single-hop} & \multicolumn{2}{c}{AdMIRaL} \\
		 & Two-hop & Overall  & Two-hop & Overall \\
		 \hline
		 BM25 & 0.065 & 0.486 & 0.370 & 0.780\\ 
		 KGAT & 0.470 & 0.955 & 0.790 & 0.968 \\ 
		 \citep{jiang-etal-2021-exploring-listwise} & 0.356 & 0.925 & 0.701 & 0.953  \\ 
		 AdMIRaL  & 0.357 & 0.928 & 0.705 & 0.956\\
		\bottomrule
	\end{tabular}}
	\caption{Document retrieval scores using various methods for retrieving the initial evidence sentences $E_1$, R: Recall@5. Note that KGAT includes gold documents before sentence re-ranking and hence has not been considered in our main experiments.}\label{tab:initial-retriever}
\end{table}

\paragraph{Sufficiency Proof with Natural Logic}
\label{section:discussion-sufficiency}
To evaluate the effectiveness of our proof-based approach to determine evidence sufficiency, we compare AdMIRaL against four baselines: i) a model that always considers the evidence to be complete/incomplete ii) a binary classifier (BART with linear head) trained to distinguish complete from incomplete input, iii) a ProofVER generated proof, iv) a Natlog proof generated solely by using lexical resources and assigning unmatched mutations the independence NatOp.  We further compare our approach against an oracle that always correctly decides whether additional evidence is required. Results on FEVER are shown in Table \ref{tab:evidence-sufficiency}. In addition to retrieval recall@5 the table shows insufficiency precision and recall for multi-hop claims. 
Considering all evidence to be insufficient is equivalent to running AdMIRaL's retriever for a fixed number of iterations $n$. AdMIRaL's sufficiency check improves notably on it, also outperforming alternative sufficiency prediction methods. However, we also note that there is substantial room for improvement as the oracle outperforms our current approach in terms of precision, which translates to a substantially higher retrieval recall@5.
 
\begin{table}[ht!]
	\centering
	\resizebox{1\linewidth}{!}{
	\begin{tabular}{l| cccc }
		\toprule
		 Model & \multicolumn{2}{c}{FEVER} & \multicolumn{2}{c}{Insufficiency}\\
		 & Two-hop & Overall & P & R  \\
		 \hline
		 All insuf &  0.69 & 0.95 & 0.59 & 1.0 \\ 
		 All suf & 0.65 & 0.94 & 0.0 & 0.0 \\ 
		 \hline
		 Classifier & 0.69 & 0.94 & 0.61 & 0.87 \\
		 ProofVER & 0.68 & 0.96 & 0.61 & 0.76 \\
		 Lexicon/KBs only & 0.69 & 0.96 & 0.60 & \textbf{0.96}\\
		 \hline 
		  Ours & \textbf{0.71} & \textbf{0.96}  & \textbf{0.70}   & 0.93 \\
		 Ours w/ Oracle Merger & 0.74 & 0.97 & 1.00 & 1.00 \\ 
		\bottomrule
	\end{tabular}}
	\caption{Document retrieval scores, R: Recall@5, EM: Exact Match. We further report scores on insufficiency prediction in terms of recall and precision.}\label{tab:evidence-sufficiency}
\end{table}

\section{Human Evaluation of Sufficiency Proofs}
\label{section:human_evlauation}
A key advantage of AdMIRaL is the added interpretability of our multi-hop retriever through our proof-based sufficiency prediction.
For instance, after an initial retrieval hop our model might predict evidence insufficiency, with the indicated span that information is missing for. If the model is not able to find the relevant information in the next hop, a user could consider a targeted modification of the claim based on the sufficiency proof to enable the model to follow a different retrieval path. Conversely, the model might erroneously indicate evidence sufficiency, so the user can supersede the model's decision in an informed manner.

To explore the interpretability of our sufficiency proofs, we conduct a forward prediction experiment \citep{doshi2017towards}. Human subjects are asked to predict whether AdMIRaL considers the evidence for a given claim to be sufficient or not, by using the generated sufficiency proof. The comparison baseline provides only the evidence sentences instead. 
Since we are evaluating the proof as an explanation mechanism to humans, we ensured that no subject was familiar with the deterministic nature of our approach. To enable non-experts to make use of the proof, we replaced the NatOps with English phrases, similar to \citep{krishna2021proofver} (see Appendix \ref{app:human-eval}).

The evaluation consists of $60$ annotations from $6$ subjects. Ten claims, each paired with an AdMIRaL NatLog proof and baseline explanation are annotated by three subjects. No subject annotates the same claim for both AdMIRaL and baseline explanation, as otherwise a subject might be influenced by the explanation it has seen before for the same claim. Using the sufficiency proofs, subjects correctly predict the model's decision in 70\% of cases, compared to the baseline's 50\%. The inter-annotator agreement for both AdMIRaL's and baseline's explanation is $0.80$ Fleiss $\kappa$ \citep{fleiss1971measuring}. Moreover, annotators predict a system's behavior using AdMIRaL's explanation 20\% faster than with the baseline, taking an average time of $51$ seconds, reduced to $24$ seconds after the first $5$ annotations.     

\section{Conclusion}
This paper explored an auto-regressive Retrieval and Rerank model for multi-hop document retrieval that is guided by a proof system based on natural logic that dynamically terminates the retrieval process if the retrieved evidence is deemed sufficient. Our model does only cause minimal memory footprint compared to current state-of-the-art retrieval models while achieving competitive retrieval recall and F$_1$. 
Human evaluation indicates that the generated proof as a sufficiency condition is interpretable, enabling a human-in-the loop in the model's retrieval process.  
Future work aims to investigate to which extent a verification model (i.e.\ ProofVER) could inform the retrieval of evidence directly, creating an end-to-end closed loop system, as well as human-in-the-loop approaches. 

\section*{Acknowledgements}
This work was supported by the Engineering and Physical Sciences Research Council Doctoral Training Partnership (EPSRC). Andreas Vlachos is supported by the ERC grant AVeriTeC (GA 865958)  and the EU H2020 grant MONITIO (GA 965576). We thank Amrith Krishna for giving us access to ProofVER, both him and Nicola De Cao for useful comments and suggestions, and Dominik Stammbach for helping reproducing their multi-hop retriever. Further, the authors would like to thank the 6 subjects who volunteered to be part of the human evaluation, namely Youmna Farag, Zhijiang Guo, Sana Kidwai, Pietro Lesci, Nedjma Ousidhoum, and Andre Schurat, as well as Christopher Bryant and Christoph Hüter for early feedback on the survey.  We finally thank the anonymous reviewers for their time and effort giving us feedback on our paper. 

\section*{Limitations}
All benchmarks explored in the paper use Wikipedia as the KB, which is homogeneous compared to heterogeneous sources professional fact-checkers use (e.g.\ news articles, encyclopedias, scientific documents). Our retrieval methods also focus solely on unstructured evidence in the form of sentences, however, as indicated, recent datasets also consider other modalities. Moreover, the datasets were constructed explicitly using hyperlinks on Wikipedia, thus our approach appears to be particularly suited to these benchmarks. However, we are not aware of a large-scale fact verification dataset that refrains from annotating data that way.
Moreover, while natural logic is interpretable, its expressiveness is limited. More complex reasoning e.g.\ involving time ranges or numbers is not suited for Natural Logic. 

\section*{Ethics Statement}
We anticipate that our retrieval system will be used in fact checking systems. Our retrieval system does not make any judgements about the truth of a statement in the real-world but only consider Wikipedia as the source of evidence to be used as the entire experimental environment has been confined to it. Wikipedia is a great collaborative resource, yet it has mistakes and noise of its own similar to any encyclopedia or knowledge source. Thus we discourage users of using our retrieval system to make absolute statements about the claims being verified, i.e. avoid using it to develop truth-tellers.

\bibliography{anthology,custom}

\begin{thebibliography}{39}
\expandafter\ifx\csname natexlab\endcsname\relax\def\natexlab#1{#1}\fi

\bibitem[{Aly et~al.(2021{\natexlab{a}})Aly, Guo, Schlichtkrull, Thorne,
  Vlachos, Christodoulopoulos, Cocarascu, and
  Mittal}]{alyFEVEROUSFactExtraction2021}
Rami Aly, Zhijiang Guo, Michael Schlichtkrull, James Thorne, Andreas Vlachos,
  Christos Christodoulopoulos, Oana Cocarascu, and Arpit Mittal.
  2021{\natexlab{a}}.
\newblock \href {https://openreview.net/forum?id=h-flVCIlstW} {{{FEVEROUS}}:
  {{Fact Extraction}} and {{VERification Over Unstructured}} and {{Structured}}
  information}.
\newblock In \emph{Thirty-fifth Conference on Neural Information Processing
  Systems Datasets and Benchmarks Track (Round 1)}.

\bibitem[{Aly et~al.(2021{\natexlab{b}})Aly, Guo, Schlichtkrull, Thorne,
  Vlachos, Christodoulopoulos, Cocarascu, and Mittal}]{aly-etal-2021-fact}
Rami Aly, Zhijiang Guo, Michael~Sejr Schlichtkrull, James Thorne, Andreas
  Vlachos, Christos Christodoulopoulos, Oana Cocarascu, and Arpit Mittal.
  2021{\natexlab{b}}.
\newblock \href {https://doi.org/10.18653/v1/2021.fever-1.1} {The fact
  extraction and {VER}ification over unstructured and structured information
  ({FEVEROUS}) shared task}.
\newblock In \emph{Proceedings of the Fourth Workshop on Fact Extraction and
  VERification (FEVER)}, pages 1--13, Dominican Republic.

\bibitem[{Asai et~al.(2020)Asai, Hashimoto, Hajishirzi, Socher, and
  Xiong}]{Asai2020LearningTR}
Akari Asai, Kazuma Hashimoto, Hannaneh Hajishirzi, Richard Socher, and Caiming
  Xiong. 2020.
\newblock Learning to retrieve reasoning paths over wikipedia graph for
  question answering.
\newblock In \emph{8th International Conference on Learning Representations,
  {{ICLR}} 2020, Addis Ababa, Ethiopia, April 26-30, 2020}.

\bibitem[{De~Cao et~al.(2021)De~Cao, Izacard, Riedel, and
  Petroni}]{decaoAutoregressiveEntityRetrieval2021}
Nicola De~Cao, Gautier Izacard, Sebastian Riedel, and Fabio Petroni. 2021.
\newblock \href {https://openreview.net/forum?id=5k8F6UU39V} {Autoregressive
  {{Entity Retrieval}}}.
\newblock In \emph{International Conference on Learning Representations}.

\bibitem[{Doshi-Velez and Kim(2017)}]{doshi2017towards}
Finale Doshi-Velez and Been Kim. 2017.
\newblock Towards a rigorous science of interpretable machine learning.
\newblock \emph{arXiv preprint arXiv:1702.08608}.

\bibitem[{Fleiss(1971)}]{fleiss1971measuring}
Joseph~L Fleiss. 1971.
\newblock Measuring nominal scale agreement among many raters.
\newblock \emph{Psychological bulletin}, 76(5):378.

\bibitem[{Graves(2018)}]{graveslucasUnderstandingPromiseLimits2018}
Lucas Graves. 2018.
\newblock Understanding the {{Promise}} and {{Limits}} of {{Automated
  Fact-Checking}}.
\newblock Technical report, Reuters Institute, University of Oxford.

\bibitem[{Guo et~al.(2022)Guo, Schlichtkrull, and
  Vlachos}]{guo-tacl-survey-2022}
Zhijiang Guo, Michael Schlichtkrull, and Andreas Vlachos. 2022.
\newblock \href {https://doi.org/10.1162/tacl_a_00454} {{A Survey on Automated
  Fact-Checking}}.
\newblock \emph{Transactions of the Association for Computational Linguistics},
  10:178--206.

\bibitem[{Hanselowski et~al.(2018)Hanselowski, Zhang, Li, Sorokin, Schiller,
  Schulz, and Gurevych}]{hanselowski-etal-2018-ukp}
Andreas Hanselowski, Hao Zhang, Zile Li, Daniil Sorokin, Benjamin Schiller,
  Claudia Schulz, and Iryna Gurevych. 2018.
\newblock \href {https://doi.org/10.18653/v1/W18-5516} {{{UKP-Athene}}:
  {{Multi-sentence}} textual entailment for claim verification}.
\newblock In \emph{Proceedings of the First Workshop on Fact Extraction and
  {{VERification}} ({{FEVER}})}, pages 103--108, {Brussels, Belgium}.

\bibitem[{Hardalov et~al.(2022)Hardalov, Arora, Nakov, and
  Augenstein}]{Hardalov2021ASO}
Momchil Hardalov, Arnav Arora, Preslav Nakov, and Isabelle Augenstein. 2022.
\newblock \href {https://doi.org/10.18653/v1/2022.findings-naacl.94} {A survey
  on stance detection for mis- and disinformation identification}.
\newblock In \emph{Findings of the Association for Computational Linguistics:
  NAACL 2022}, pages 1259--1277, Seattle, United States.

\bibitem[{Hidey et~al.(2020)Hidey, Chakrabarty, Alhindi, Varia, Krstovski,
  Diab, and Muresan}]{Hidey2020DeSePtionDS}
Christopher Hidey, Tuhin Chakrabarty, Tariq Alhindi, Siddharth Varia, Kriste
  Krstovski, Mona Diab, and Smaranda Muresan. 2020.
\newblock \href {https://doi.org/10.18653/v1/2020.acl-main.761} {{{DeSePtion}}:
  {{Dual}} sequence prediction and adversarial examples for improved
  fact-checking}.
\newblock In \emph{Proceedings of the 58th Annual Meeting of the Association
  for Computational Linguistics}, pages 8593--8606, {Online}. {Association for
  Computational Linguistics}.

\bibitem[{Jiang et~al.(2021{\natexlab{a}})Jiang, Pradeep, and
  Lin}]{jiang-etal-2021-exploring-listwise}
Kelvin Jiang, Ronak Pradeep, and Jimmy Lin. 2021{\natexlab{a}}.
\newblock \href {https://doi.org/10.18653/v1/2021.acl-short.51} {Exploring
  listwise evidence reasoning with t5 for fact verification}.
\newblock In \emph{Proceedings of the 59th Annual Meeting of the Association
  for Computational Linguistics and the 11th International Joint Conference on
  Natural Language Processing (Volume 2: Short Papers)}, pages 402--410,
  Online.

\bibitem[{Jiang et~al.(2021{\natexlab{b}})Jiang, Pradeep, and
  Lin}]{jiangExploringListwiseEvidence2021}
Kelvin Jiang, Ronak Pradeep, and Jimmy Lin. 2021{\natexlab{b}}.
\newblock \href {https://doi.org/10.18653/v1/2021.acl-short.51} {Exploring
  {{Listwise Evidence Reasoning}} with {{T5}} for {{Fact Verification}}}.
\newblock In \emph{Proceedings of the 59th {{Annual Meeting}} of the
  {{Association}} for {{Computational Linguistics}} and the 11th
  {{International Joint Conference}} on {{Natural Language Processing}}
  ({{Volume}} 2: {{Short Papers}})}, pages 402--410, {Online}.

\bibitem[{Jiang et~al.(2020)Jiang, Bordia, Zhong, Dognin, Singh, and
  Bansal}]{jiang2020hover}
Yichen Jiang, Shikha Bordia, Zheng Zhong, Charles Dognin, Maneesh Singh, and
  Mohit Bansal. 2020.
\newblock \href {https://doi.org/10.18653/v1/2020.findings-emnlp.309}
  {{{HoVer}}: {{A}} dataset for many-hop fact extraction and claim
  verification}.
\newblock In \emph{Findings of the Association for Computational Linguistics:
  {{EMNLP}} 2020}, pages 3441--3460, {Online}.

\bibitem[{Karpukhin et~al.(2020)Karpukhin, Oguz, Min, Lewis, Wu, Edunov, Chen,
  and Yih}]{Karpukhin2020DensePR}
Vladimir Karpukhin, Barlas Oguz, Sewon Min, Patrick Lewis, Ledell Wu, Sergey
  Edunov, Danqi Chen, and Wen-tau Yih. 2020.
\newblock \href {https://doi.org/10.18653/v1/2020.emnlp-main.550} {Dense
  passage retrieval for open-domain question answering}.
\newblock In \emph{Proceedings of the 2020 Conference on Empirical Methods in
  Natural Language Processing ({{EMNLP}})}, pages 6769--6781, {Online}.

\bibitem[{Khattab et~al.(2021)Khattab, Potts, and
  Zaharia}]{khattabBaleenRobustMultiHop2021}
Omar Khattab, Christopher Potts, and Matei Zaharia. 2021.
\newblock \href {https://openreview.net/forum?id=Ghk0AJ8XtVx} {Baleen: Robust
  multi-hop reasoning at scale via condensed retrieval}.
\newblock In \emph{Advances in Neural Information Processing Systems}, online.

\bibitem[{Krishna et~al.(2022)Krishna, Riedel, and
  Vlachos}]{krishna2021proofver}
Amrith Krishna, Sebastian Riedel, and Andreas Vlachos. 2022.
\newblock \href {https://doi.org/10.1162/tacl_a_00503} {{ProoFVer: Natural
  Logic Theorem Proving for Fact Verification}}.
\newblock \emph{Transactions of the Association for Computational Linguistics},
  10:1013--1030.

\bibitem[{Lewis et~al.(2020)Lewis, Liu, Goyal, Ghazvininejad, Mohamed, Levy,
  Stoyanov, and Zettlemoyer}]{lewis-etal-2020-bart}
Mike Lewis, Yinhan Liu, Naman Goyal, Marjan Ghazvininejad, Abdelrahman Mohamed,
  Omer Levy, Veselin Stoyanov, and Luke Zettlemoyer. 2020.
\newblock \href {https://doi.org/10.18653/v1/2020.acl-main.703} {{BART}:
  Denoising sequence-to-sequence pre-training for natural language generation,
  translation, and comprehension}.
\newblock In \emph{Proceedings of the 58th Annual Meeting of the Association
  for Computational Linguistics}, pages 7871--7880, Online.

\bibitem[{Li et~al.(2021)Li, Li, Shang, Jiang, Liu, Sun, Ji, and
  Liu}]{li2021hopretriever}
Shaobo Li, Xiaoguang Li, Lifeng Shang, Xin Jiang, Qun Liu, Chengjie Sun,
  Zhenzhou Ji, and Bingquan Liu. 2021.
\newblock Hopretriever: Retrieve hops over wikipedia to answer complex
  questions.
\newblock In \emph{Proceedings of the AAAI Conference on Artificial
  Intelligence}, volume~35, pages 13279--13287.

\bibitem[{Lin et~al.(2021)Lin, Ma, Lin, Yang, Pradeep, and
  Nogueira}]{lin2021pyserini}
Jimmy Lin, Xueguang Ma, Sheng-Chieh Lin, Jheng-Hong Yang, Ronak Pradeep, and
  Rodrigo Nogueira. 2021.
\newblock \href {https://dl.acm.org/doi/10.1145/3404835.3463238} {Pyserini: An
  easy-to-use python toolkit to support replicable ir research with sparse and
  dense representations}.
\newblock In \emph{Proceedings of the 44th International ACM SIGIR Conference
  on Research and Development in Information Retrieval (SIGIR '21).}, online.

\bibitem[{Liu et~al.(2020)Liu, Xiong, Sun, and
  Liu}]{liuFinegrainedFactVerification2020}
Zhenghao Liu, Chenyan Xiong, Maosong Sun, and Zhiyuan Liu. 2020.
\newblock \href {https://doi.org/10.18653/v1/2020.acl-main.655} {Fine-grained
  {{Fact Verification}} with {{Kernel Graph Attention Network}}}.
\newblock In \emph{Proceedings of the 58th {{Annual Meeting}} of the
  {{Association}} for {{Computational Linguistics}}}, pages 7342--7351,
  {Online}.

\bibitem[{MacCartney and Manning(2014)}]{maccartney2014natural}
Bill MacCartney and Christopher~D Manning. 2014.
\newblock Natural logic and natural language inference.
\newblock In \emph{Computing meaning}, pages 129--147. Springer.

\bibitem[{Malon(2021)}]{malon-2021-team}
Christopher Malon. 2021.
\newblock \href {https://doi.org/10.18653/v1/2021.fever-1.5} {Team {Papelo} at
  {FEVEROUS}: Multi-hop evidence pursuit}.
\newblock In \emph{Proceedings of the Fourth Workshop on Fact Extraction and
  VERification (FEVER)}, pages 40--49, Dominican Republic. Association for
  Computational Linguistics.

\bibitem[{Miller(1995)}]{miller-1995-wordnet}
George~A. Miller. 1995.
\newblock \href {https://doi.org/10.1145/219717.219748} {Wordnet: A lexical
  database for english}.
\newblock \emph{Commun. ACM}, 38(11):39–41.

\bibitem[{Nie et~al.(2019)Nie, Chen, and Bansal}]{Nie2019CombiningFE}
Yixin Nie, Haonan Chen, and Mohit Bansal. 2019.
\newblock \href {https://doi.org/10.1609/aaai.v33i01.33016859} {Combining fact
  extraction and verification with neural semantic matching networks}.
\newblock In \emph{The Thirty-Third {{AAAI}} Conference on Artificial
  Intelligence, {{AAAI}} 2019, the Thirty-First Innovative Applications of
  Artificial Intelligence Conference, {{IAAI}} 2019, the Ninth {{AAAI}}
  Symposium on Educational Advances in Artificial Intelligence, {{EAAI}} 2019,
  Honolulu, Hawaii, {{USA}}, January 27 - February 1, 2019}, pages 6859--6866.

\bibitem[{Paszke et~al.(2019)Paszke, Gross, Massa, Lerer, Bradbury, Chanan,
  Killeen, Lin, Gimelshein, Antiga et~al.}]{paszke2019pytorch}
Adam Paszke, Sam Gross, Francisco Massa, Adam Lerer, James Bradbury, Gregory
  Chanan, Trevor Killeen, Zeming Lin, Natalia Gimelshein, Luca Antiga, et~al.
  2019.
\newblock Pytorch: An imperative style, high-performance deep learning library.
\newblock \emph{Advances in neural information processing systems}, 32.

\bibitem[{Pavlick et~al.(2015)Pavlick, Rastogi, Ganitkevitch, Van~Durme, and
  Callison-Burch}]{pavlick-etal-2015-ppdb}
Ellie Pavlick, Pushpendre Rastogi, Juri Ganitkevitch, Benjamin Van~Durme, and
  Chris Callison-Burch. 2015.
\newblock \href {https://doi.org/10.3115/v1/P15-2070} {{PPDB} 2.0: Better
  paraphrase ranking, fine-grained entailment relations, word embeddings, and
  style classification}.
\newblock In \emph{Proceedings of the 53rd Annual Meeting of the Association
  for Computational Linguistics and the 7th International Joint Conference on
  Natural Language Processing (Volume 2: Short Papers)}, pages 425--430,
  Beijing, China.

\bibitem[{Petroni et~al.(2021)Petroni, Piktus, Fan, Lewis, Yazdani, De~Cao,
  Thorne, Jernite, Karpukhin, Maillard, Plachouras, Rockt{\"a}schel, and
  Riedel}]{petroni-etal-2021-kilt}
Fabio Petroni, Aleksandra Piktus, Angela Fan, Patrick Lewis, Majid Yazdani,
  Nicola De~Cao, James Thorne, Yacine Jernite, Vladimir Karpukhin, Jean
  Maillard, Vassilis Plachouras, Tim Rockt{\"a}schel, and Sebastian Riedel.
  2021.
\newblock \href {https://doi.org/10.18653/v1/2021.naacl-main.200} {{KILT}: a
  benchmark for knowledge intensive language tasks}.
\newblock In \emph{Proceedings of the 2021 Conference of the North American
  Chapter of the Association for Computational Linguistics: Human Language
  Technologies}, pages 2523--2544, Online.

\bibitem[{Petroni et~al.(2019)Petroni, Rockt{\"a}schel, Riedel, Lewis, Bakhtin,
  Wu, and Miller}]{petroni-etal-2019-language}
Fabio Petroni, Tim Rockt{\"a}schel, Sebastian Riedel, Patrick Lewis, Anton
  Bakhtin, Yuxiang Wu, and Alexander Miller. 2019.
\newblock \href {https://doi.org/10.18653/v1/D19-1250} {Language models as
  knowledge bases?}
\newblock In \emph{Proceedings of the 2019 Conference on Empirical Methods in
  Natural Language Processing and the 9th International Joint Conference on
  Natural Language Processing ({{EMNLP-IJCNLP}})}, pages 2463--2473, {Hong
  Kong, China}.

\bibitem[{Qi et~al.(2021)Qi, Lee, Sido, and Manning}]{qi-etal-2021-answering}
Peng Qi, Haejun Lee, Tg~Sido, and Christopher Manning. 2021.
\newblock \href {https://doi.org/10.18653/v1/2021.emnlp-main.292} {Answering
  open-domain questions of varying reasoning steps from text}.
\newblock In \emph{Proceedings of the 2021 Conference on Empirical Methods in
  Natural Language Processing}, pages 3599--3614, Online and Punta Cana,
  Dominican Republic.

\bibitem[{Radford et~al.(2019)Radford, Wu, Child, Luan, Amodei, and
  Sutskever}]{radford2019language}
Alec Radford, Jeffrey Wu, Rewon Child, David Luan, Dario Amodei, and Ilya
  Sutskever. 2019.
\newblock Language models are unsupervised multitask learners.
\newblock \emph{OpenAI blog}, 1(8):9.

\bibitem[{Roth and Schulte~im
  Walde(2014)}]{roth-schulte-im-walde-2014-combining}
Michael Roth and Sabine Schulte~im Walde. 2014.
\newblock \href {https://doi.org/10.3115/v1/P14-2086} {Combining word patterns
  and discourse markers for paradigmatic relation classification}.
\newblock In \emph{Proceedings of the 52nd Annual Meeting of the Association
  for Computational Linguistics (Volume 2: Short Papers)}, pages 524--530,
  Baltimore, Maryland.

\bibitem[{Stammbach(2021)}]{stammbach-2021-evidence}
Dominik Stammbach. 2021.
\newblock \href {https://doi.org/10.18653/v1/2021.fever-1.2} {Evidence
  selection as a token-level prediction task}.
\newblock In \emph{Proceedings of the Fourth Workshop on Fact Extraction and
  VERification (FEVER)}, pages 14--20, Dominican Republic.

\bibitem[{Stammbach and Neumann(2019)}]{stammbach-neumann-2019-team}
Dominik Stammbach and Guenter Neumann. 2019.
\newblock \href {https://doi.org/10.18653/v1/D19-6616} {Team {{DOMLIN}}:
  {{Exploiting}} evidence enhancement for the {{FEVER}} shared task}.
\newblock In \emph{Proceedings of the Second Workshop on Fact Extraction and
  {{VERification}} ({{FEVER}})}, pages 105--109, {Hong Kong, China}.

\bibitem[{Thorne et~al.(2018)Thorne, Vlachos, Christodoulopoulos, and
  Mittal}]{thorneFEVERLargescaleDataset2018}
James Thorne, Andreas Vlachos, Christos Christodoulopoulos, and Arpit Mittal.
  2018.
\newblock \href {https://doi.org/10.18653/v1/N18-1074} {{{FEVER}}: A
  {{Large-scale Dataset}} for {{Fact Extraction}} and {{VERification}}}.
\newblock In \emph{Proceedings of the 2018 {{Conference}} of the {{North
  American Chapter}} of the {{Association}} for {{Computational Linguistics}}:
  {{Human Language Technologies}}}, pages 809--819, {New Orleans, Louisiana}.

\bibitem[{Thorne et~al.(2019)Thorne, Vlachos, Cocarascu, Christodoulopoulos,
  and Mittal}]{thorne-etal-2019-fever2}
James Thorne, Andreas Vlachos, Oana Cocarascu, Christos Christodoulopoulos, and
  Arpit Mittal. 2019.
\newblock \href {https://doi.org/10.18653/v1/D19-6601} {The {FEVER}2.0 shared
  task}.
\newblock In \emph{Proceedings of the Second Workshop on Fact Extraction and
  VERification (FEVER)}, pages 1--6, Hong Kong, China.

\bibitem[{Wolf et~al.(2020)Wolf, Debut, Sanh, Chaumond, Delangue, Moi, Cistac,
  Rault, Louf, Funtowicz, Davison, Shleifer, von Platen, Ma, Jernite, Plu, Xu,
  Le~Scao, Gugger, Drame, Lhoest, and Rush}]{wolf2019huggingface}
Thomas Wolf, Lysandre Debut, Victor Sanh, Julien Chaumond, Clement Delangue,
  Anthony Moi, Pierric Cistac, Tim Rault, Remi Louf, Morgan Funtowicz, Joe
  Davison, Sam Shleifer, Patrick von Platen, Clara Ma, Yacine Jernite, Julien
  Plu, Canwen Xu, Teven Le~Scao, Sylvain Gugger, Mariama Drame, Quentin Lhoest,
  and Alexander Rush. 2020.
\newblock \href {https://doi.org/10.18653/v1/2020.emnlp-demos.6} {Transformers:
  State-of-the-art natural language processing}.
\newblock In \emph{Proceedings of the 2020 Conference on Empirical Methods in
  Natural Language Processing: System Demonstrations}, pages 38--45, Online.

\bibitem[{Xiong et~al.(2021)Xiong, Li, Iyer, Du, Lewis, Wang, Mehdad, Yih,
  Riedel, Kiela, and Oguz}]{xiongAnsweringComplexOpenDomain2020}
Wenhan Xiong, Xiang Li, Srini Iyer, Jingfei Du, Patrick Lewis, William~Yang
  Wang, Yashar Mehdad, Scott Yih, Sebastian Riedel, Douwe Kiela, and Barlas
  Oguz. 2021.
\newblock \href {https://openreview.net/forum?id=EMHoBG0avc1} {Answering
  complex open-domain questions with multi-hop dense retrieval}.
\newblock In \emph{International Conference on Learning Representations},
  online.

\bibitem[{Zhong et~al.(2020)Zhong, Xu, Tang, Xu, Duan, Zhou, Wang, and
  Yin}]{Zhong2020ReasoningOS}
Wanjun Zhong, Jingjing Xu, Duyu Tang, Zenan Xu, Nan Duan, Ming Zhou, Jiahai
  Wang, and Jian Yin. 2020.
\newblock \href {https://doi.org/10.18653/v1/2020.acl-main.549} {Reasoning over
  semantic-level graph for fact checking}.
\newblock In \emph{Proceedings of the 58th Annual Meeting of the Association
  for Computational Linguistics}, pages 6170--6180, {Online}.

\end{thebibliography}
\bibliographystyle{acl_natbib}

\appendix

\section{Appendix}
\label{sec:appendix}

\subsection{Autoregressive Document Retrieval}
\label{app:autoregress}

\begin{table*}[ht!]
	\centering
	\resizebox{0.80\linewidth}{!}{
	\begin{tabular}{ll| cc cc cc}
		\toprule
	&	Model/Datasets & \multicolumn{2}{c}{FEVER} & \multicolumn{2}{c}{FEVEROUS-S} & \multicolumn{2}{c}{HoVer}\\
	& & 2-hop & Overall & 2-hop & Overall & 2-hop & Overall \\
		\midrule
	\multirow{ 3}{*}{Single-Hop} &	BM25  & 0.101  & 0.367 &  0.327 & 0.548 & 0.385 & 0.141 \\
	 & GENRE & 0.195  & 0.609 & 0.330 & 0.0.367 & 0.396 & 0.152 \\
	 & AdMIRaL-1hop  & 0.359 & 0.643 & 0.456 & 0.612 & 0.653 & 0.289  \\
	 \hline
	\multirow{ 2}{*}{Multi-Hop}
	& Hyperlinks & 0.412 & \underline{0.647} & \underline{0.488} & \underline{0.617} & \underline{0.711} & \underline{0.441} \\
	&	MDR$^\dagger$ & \underline{0.550} & -- & --&-- & -- & -- \\
		\hline
	&	AdMIRaL (Ours) & \textbf{0.596} & \textbf{0.667} & \textbf{0.579} & \textbf{0.634} & \textbf{0.783}  & \textbf{0.559} \\
		\bottomrule
	\end{tabular}}
	\caption{F$_1$ document retrieval scores for 2-hop, and overall scores. 
	To compare with previous work on HoVer, we report recall@100 for \emph{supported} claims on dev. $\dagger$ indicate results taken from \citet{xiongAnsweringComplexOpenDomain2020}. Results from  \citet{khattabBaleenRobustMultiHop2021} excluded as computation of $F_1$ in HoVer is unclear and script not accessible. \textbf{Bold} numbers indicate best and \underline{underline} the second-best score.
	}\label{tab:overview-document-retrieval-F1}
\end{table*}

\paragraph{Generative Model} We use a pre-trained seq2seq model, namely BART \cite{lewis-etal-2020-bart}, which allows the model to capture both surface-level information and semantic aspects between the claim and candidate sentences using cross-attention in its encoder while its decoder is attending over the hidden sequence during generation. The input is structured so that the claim is followed by the sentences, each separated by end-of-sentence tokens: c </s> $e_0$, </s> ... </s> $e_k$. Each evidence sentence in $E_i$ is preceded by the corresponding document title in square brackets, e.g. ``\emph{[James McBrayer] Jack McBrayer (born May 27... [Tom Bergeron] Tom Bergeron (born May 6, 1955) ...}". For decoding the sentence identifier are generated first, followed by the document identifier. For the dynamic markup decoding we use square brackets to swap between search spaces while each sentence identifier is separated by a space: $e_{p1} e_{p2} [ d_p ]$. The input to the proof sufficiency module is formatted according to the requirements of ProofVER. Thy dynamic markup of the proof's output is structured by using braces to surround a claim span, square brackets to cover the evidence span, and the token after the closing square bracket to be the natural logic operator.

\paragraph{Training} We train a separate model for each hop $t$ using maximum likelihood estimation, 
following the Neural Machine Translation fine-tuning of BART \cite{lewis-etal-2020-bart}. Given an ordered list of gold evidence sentences E$_g$, we generate training data by considering only samples with gold evidence sentences equal to the number of hops $t$.  Since the model is trained as a pointwise reranker, we keep the top $t-1$ gold evidence sentences in the input, i.e.\ in $E_t$. and the output is subsequently comprised of a single sequence $p\in P_t$. We generate $p$ as the output label by concatenating the sentence ids of the $t-1$ gold evidence sentences with the document title the remaining gold sentence is representing. Since we consider the top $l$ sentences during inference, with $l \geq t$, we further sample $l-t$ negative sentences from the document candidates $D_t$ and add them to the input. 
We further shuffle the input randomly, forcing the model to learn to attend to all input sentences. For instance, given the example in Figure \ref{fig:feverous_example}, the training data for hop $t=2$ would contain as input the claim, ``\emph{Seth Meyers hosted the ceremony}", and negative samples, and the output label could be \emph{E2 [ Seth Meyers ]}, with $2$ varying depending on its position in the input.
In cases where an explicit ordering of evidence sentences is not available, we generate all $t-1$ possible ways of keeping $t-1$ gold sentences in the input and using the other as the output.

\subsection{Implementation Details}
\label{app:impl}

All models are implemented using PyTorch \citep{paszke2019pytorch}. The autoregressive model for both retrieval and proof generation are based on the \texttt{Huggingface} \citep{wolf2019huggingface} implementation of BART \citep{lewis-etal-2020-bart} and GENRE \citep{decaoAutoregressiveEntityRetrieval2021}. For all experiments we use a beam size of $25$ for the autoregressive generation, and a beam size of $5$ for the generation of the sufficiency proof. We used default hyperparameters of BART on all experiments. In case $D_t$ contains less documents than considered by the metric (e.g. recall@5 but number of documents $k<5$) we add additional documents from $D_{t-1}$. All experiments were run on a machine with a \emph{single} Quadro RTX 8000 and 64GB RAM memory.
\citet{krishna2021proofver} kindly provided us access to their ProofVER model. For BM25 we set $k1=0.6$ and $b=0.4$, following recommendations of Pyserini.

\subsection{Further Results}
\label{app:further-results}
Table \ref{tab:overview-document-retrieval-F1} shows results F$_1$ scores on FEVER, FEVEROUS-S, and HoVer.

\subsection{Human Evaluation}
All subjects in the human evaluation are undergraduate/graduate/postgraduates students in either computer science or linguistics. 4 subjects are male, 2 female. None of the subjects had prior knowledge on natural language inference.
\label{app:human-eval}

\begin{table}[ht!]
\resizebox{\linewidth}{!}{
	\begin{tabular}{l| c }
	\toprule
	    NatOP & Paraphrase \\
	    \midrule
		$\equiv$ & Equivalent Spans\\
		$\neg$ & Evidence span refutes claim span \\
		$\downharpoonleft \! \upharpoonright$ & Evidence span contradicts the claim span \\
		\# & Unrelated claim span and evidence span\\
	\bottomrule
	\end{tabular}}
	\caption{NatOPs and their corresponding paraphrases.}
\end{table}

\end{document}